%% file: Manuscript.tex
\documentclass[letterpaper]{article} 
\usepackage{aaai2026}  
\usepackage{times}  
\usepackage{helvet}  
\usepackage{courier}  
\usepackage[hyphens]{url}  
\usepackage{graphicx} 
\usepackage{natbib}  
\usepackage{caption} 
\usepackage{tabularx} 
\usepackage{subcaption}
\usepackage{algorithm}
\usepackage{amsmath}
\usepackage{algorithmic}

\usepackage{newfloat}
\usepackage{listings}
\DeclareCaptionStyle{ruled}{labelfont=normalfont,labelsep=colon,strut=off} 
\floatstyle{ruled}
\newfloat{listing}{tb}{lst}{}
\floatname{listing}{Listing}
\title{Beyond RGB: Leveraging Vision Transformers for Thermal Weapon Segmentation}
\author{
    Akhila Kambhatla \textsuperscript{\rm 1}, Ahmed R Khaled \textsuperscript{\rm 2} \\
}
\affiliations{
    \textsuperscript{\rm 1}School of Computing, SIUC, akhila.kambhatla@siu.edu\\
    \textsuperscript{\rm 2} School of Computing, SIUC. khaled.ahmed@siu.edu

    1230 Lincoln Dr, Carbondale, IL 62901.\\

}

\usepackage{bibentry}

\begin{document}

\maketitle

\begin{abstract}
 Thermal weapon segmentation is crucial for surveillance and security applications, enabling robust detection under low-light and visually obscured conditions where RGB-based systems fail. While convolutional neural networks (CNNs) dominate thermal segmentation literature, their ability to capture long-range dependencies and fine structural details is limited. Vision Transformers (ViTs), with their global context modeling capabilities, have achieved state-of-the-art results in RGB segmentation tasks, yet their potential in thermal weapon segmentation remains underexplored. This work adapts and evaluates four transformer-based architectures—SegFormer, DeepLabV3+, SegNeXt, and Swin Transformer—for binary weapon segmentation on a custom thermal dataset comprising 9,711 images collected from real-world surveillance videos and automatically annotated using SAM2. We employ standard augmentation strategies within the MMSegmentation framework to ensure robust model training and fair architectural comparison. Experimental results demonstrate significant improvements in segmentation performance: SegFormer-b5 achieves the highest mIoU (94.15\%) and Pixel Accuracy (97.04\%), while SegFormer-b0 provides the fastest inference speed (98.32 FPS) with competitive mIoU (90.84\%). SegNeXt-mscan\_s offers balanced performance with 85.12 FPS and 92.24\% mIoU, and DeepLabV3+ R101-D8 reaches 92.76\% mIoU at 29.86 FPS. The transformer architectures demonstrate robust generalization capabilities for weapon detection in low-light and occluded thermal environments, with flexible accuracy-speed trade-offs suitable for diverse real-time security applications.
\end{abstract}


\section{Introduction}
Surveillance systems that rely on RGB (visible-light) cameras often face significant hurdles—especially at night or in poorly lit conditions. These cameras struggle with low signal-to-noise ratios, motion blur, and overall poor image quality when there's little ambient light or sudden changes in illumination \cite{intro-1}. Additionally, clever camouflage or complex backgrounds can easily fool RGB-based detectors. By contrast, thermal imaging captures heat radiation directly emitted by objects, making it inherently robust to variations in lighting, fog, smoke, or visual clutte\cite{intro-2}.Thermal imaging offers an elegant solution: by sensing thermal radiation rather than visible light, it remains effective in darkness or through obscuration—making it ideal for challenging surveillance environments. Thermal segmentation methods are already being explored for their robustness in broad applications, yet comprehensive, domain-specific evaluations remain sparse\cite{intro-7}. 

Weapon detection in thermal imagery is especially promising because firearms tend to show distinct heat signatures, whether they're obscured by clothing or concealed in darkness\cite{intro-2, intro-3,intro-5, intro-6}. Thermal-based detection is invaluable for high-stakes scenarios such as airport security, public transportation hubs, critical infrastructure surveillance, and military operations, where lighting cannot be guaranteed and visual stealth is often employed\cite{intro-4, intro-3}. It’s not just about detection; reducing false positives by focusing on the physical heat patterns of weapons provides both accuracy and efficiency in real-world security workflows.

Despite growing interest, research on thermal-based weapon segmentation remains limited and largely reliant on CNN-based models. Most prior research relies on CNN-based approaches—such as variations of U-Net, Mask R‑CNN, or YOLO—trained on thermal or fused multi-modal data. For example, one study combined fine-tuned VGG-19 and YOLO-V3 architectures to detect concealed pistols in thermal imagery, achieving high F1-scores and real-time inference. However, it uses bounding boxes rather than pixel-level segmentation\cite{intro-9,intro-10,intro-11}. Meanwhile, transformer-based approaches—which excel in global context modeling—have transformed segmentation tasks in the RGB domain. Yet their effectiveness in thermal weapon segmentation has not been systematically assessed. Transformer-based methods, which have reshaped RGB-based segmentation with their global-context modeling capabilities, have yet to be thoroughly investigated for thermal weapon segmentation\cite{intro-12}. 

Our work addresses this gap. \textbf{Can vision transformer-based architectures outperform conventional CNN models in the binary segmentation of weapons in thermal imagery, while maintaining a suitable trade-off between accuracy and real-time performance?} We hypothesize that transformer-based architectures, due to their superior global context modeling and long-range dependency capture, will outperform CNN baselines in thermal weapon segmentation accuracy, offering flexible accuracy-speed trade-offs across different model configurations.

Summarizing the gaps: (a) insufficient exploration of transformers in thermal segmentation, (b) thermal datasets are small and lack diversity, and (c) real-world deployment requires both high accuracy and low latency. To bridge these, our study offers:
\begin{itemize}
    \item A \textbf{comparative evaluation} of multiple transformer architectures (SegFormer, DeepLabV3+, SegNeXt, Swin Transformer) on thermal weapon data.
    \item \textbf{Unique \& Real-time} thermal videos are captured using FLIR GF77 OGI camera, autolabelled using SAM2 (Segment Anything Model), and trained by transformers for real-time firearm segmentation. 
    \item \textbf{Domain-specific adaptations}, including thermal preprocessing and aggressive augmentation, to maximize learning from limited data.
    \item A \textbf{balanced analysis} of segmentation accuracy versus inference speed, guiding the choice of architecture depending on deployment needs.
\end{itemize}
We will expand on these innovations and validate them with experiments and results in the following Methodology and Results sections.

\section{2. Literature Review}
\subsection{2.1 Early Thermal Imaging Techniques in Security Applications via CNN's}
Thermal imaging has long been leveraged in surveillance systems for human and object detection, particularly in low-light or visually complex environments\cite{kambhatla-1,lonewolf}. Early approaches relied heavily on traditional segmentation methods such as thresholding, background subtraction, and edge detection for candidate extraction. These methods, while computationally lightweight, are highly sensitive to noise and environmental changes, limiting their robustness in real-world scenarios\cite{rw-4, rw-3}. 

With the advent of deep learning, CNN-based architectures have substantially improved segmentation performance in thermal domains. Work focusing on thermal firearm detection has yielded strong results using CNN-based models. For instance, \cite{intro-7} developed a hybrid deep learning pipeline using VGG‑19 for classification and YOLO‑V3 for localization, achieving an F1 score of 0.84 and real-time detection (\~10ms) of concealed pistols in thermal imagery.
The U-Net architecture \cite{unet} has become a staple in pixel-level prediction tasks due to its encoder–decoder design with skip connections, enabling the preservation of spatial detail. In thermal imagery, U-Net and ACF variants have been applied for pedestrian segmentation in night surveillance systems \cite{rw-5}.

Mask R-CNN \cite{mask}, a two-stage detection-and-segmentation framework, has been adapted to thermal datasets for human detection. For instance, a ResNet50-based Mask R-CNN achieved an F-score of 87.85\%, recall of 79.33\%, and precision of 98.41\% on FLIR thermal pedestrian datasets \cite{tan}. Similarly, DeepLabV3+ \cite{chen}, with its Atrous Spatial Pyramid Pooling (ASPP) for multi-scale context capture, has appeared frequently in thermal segmentation surveys \cite{ss-1}. Building on this, study presented a two-stage thermal handgun detection system that reduces false positives by verifying weapon-person associations, implemented on wearable devices with a custom thermal dataset, delivering an mAP@50‑95 of 64.52\%.
\subsection{2.2 Scope of Semantic Segmentation}
Beyond weapons, the segmentation of thermal images has been explored across general contexts. A comparative survey of semantic segmentation in thermal imagery outlines the challenges of thermal crossover, low resolution, and dataset scarcity, urging domain-specific approaches\cite{rw-2,rw-6}. In a complementary study, \cite{intro-9} reviewed infrared-based segmentation methods, emphasizing that, despite multiple use cases—from agriculture to defense—few methods focus solely on thermal input. Additionally, \cite{rw-6} introduced EC‑CNN, a gated edge‑conditioned CNN that incorporates edge priors into thermal segmentation, validated on the novel SODA dataset encompassing over 7,000 annotated thermal images.

While CNNs dominate thermal segmentation literature, Vision Transformers (ViTs) have recently redefined semantic segmentation in the RGB domain by capturing long-range dependencies via self-attention \cite{vit}. Architectures such as SegFormer \cite{segformer}, SegNeXt \cite{segnext}, and transformer-enhanced DeepLabV3+ demonstrate state-of-the-art results in natural image segmentation tasks but have not yet been comprehensively explored for thermal imagery.

Transformer-based adaptations of U-Net are particularly relevant in cross-domain contexts. Swin-Unet \cite{swinunet}, which integrates the Swin Transformer as both encoder and decoder, has shown strong results in medical segmentation tasks, outperforming CNN baselines. Similarly, DS-TransUNet \cite{transunet} employs a dual-scale Swin Transformer in a U-Net-like framework for multi-scale feature extraction, delivering superior accuracy in volumetric segmentation benchmarks.

In thermal-like modalities, CSI-Net \cite{csi-net} combines CNN feature extraction with a Swin Transformer backbone for small-target detection in infrared imagery, yielding marked gains in mIoU and detection probability. Importantly, survey on semantic segmentation for thermal images \cite{ss-2} highlights challenges such as low spatial resolution and ambiguous object boundaries, and emphasizes that transformer-based methods remain largely underexplored in this field.
\input{datasets__tab}
\subsection{2.3 Vision Transformers for Segmentation}
The arrival of Vision Transformers (ViTs) marked a paradigm shift in dense visual tasks. Li et al. demonstrate that ViT-based segmentation outperforms CNN counterparts owing to full-image self-attention and global modeling capabilities \cite{ss-3}.  Several comprehensive surveys outline these advancements: distills transformer-based segmentation methods within a unified meta-architecture \cite{ss-4}. A broader comparative survey highlights progress across vision tasks—including segmentation—with transformers displaying strong performance across modalities \cite{ss-2}.
\subsection{2.4 Security-focused thermal Detection Studies}
Thermal-based weapon detection has received growing attention in recent years. A two-stage concealed handgun detection framework using thermal imaging and deep learning achieved a mAP@50–95 of 64.52\% on a custom dataset, optimized for embedded devices and low-end hardware (\cite{rw-7}. Another study applied YOLOv8 for object segmentation in thermal imagery, reporting a mAP50 of 82\% and mAP50–90 of 59.3\% on a dataset of 1,898 images, underscoring the feasibility of real-time detection for security applications \cite{Sharma}. Beyond object detection, a recent review on thermal heat-map–based weapon detection \cite{heatmap} suggests that combining thermal imagery, deep learning, and visualization techniques can improve situational awareness in surveillance systems. However, these studies primarily focus on bounding-box detection rather than fine-grained pixel-level segmentation.
\subsection{2.5 Synthesis of our research: Gaps \& Focus}
Thermal weapon detection research suffers from small, proprietary datasets ($<$ 2,000 images) with inconsistent annotations, as seen in Table[\ref{datasets_res}]. This scarcity motivated our 9,711-image dataset with SAM2-assisted pixel-level annotations, addressing critical gaps in scale, quality, and public availability. To the best of our knowledge, no segmentation work exists for weapons using Vision Transformers.
\subsubsection{Gaps}
\begin{itemize}
    \item CNN-based thermal weapon detection shows strong real-time capability but primarily focuses on bounding-box localization or coarse segmentation.
    \item Thermal segmentation research signals both the domain’s challenges (data scarcity, low visual detail) and promising methods (edge priors, multimodal fusion), yet remains limited in weapon-specific contexts.
    \item Vision Transformers offer transformative potential via global context modeling and simplified architectures, but have not yet been applied to thermal weapon segmentation.
\end{itemize}
\subsubsection{Research Synthesis}
\begin{itemize}
    \item A first-of-its-kind evaluation of transformer-based architectures—SegFormer, DeepLabV3+, SegNeXt, and Swin Transformer—on thermal weapon segmentation.
    \item Domain-specific enhancements, including tailored preprocessing and augmentation to mitigate thermal imagery limitations.
    \item A balanced evaluation comparing segmentation accuracy and inference latency, essential for practical deployment. 
    \item The current dataset collected from real-time thermal videos are captured using an FLIR GF77 OGI camera under varying conditions and automated by SAM2. It comprises 50\% of rifle class, 20\% of handguns, 20\% of revolver class and remaining 10\% as human class. Rainbow high contrast palette is chosen for contour and edge detection of the weapon. 
\end{itemize}

\section{3. Methodology}
\subsection{3.1 Dataset Description \& PreProcessing}
Our experiments use a custom thermal weapon segmentation dataset for security applications with high-resolution thermal images annotated at pixel level using binary labels: weapon and background. The dataset was collected from real-time thermal videos captured using an \textbf{FLIR GF77 OGI} camera under varying conditions and automated by SAM2 for operational security relevance. Video streams were sampled at fixed intervals to extract frames with varied backgrounds, object scales, and environmental conditions, including different temperatures, distances, and occlusion scenarios, ensuring real-world robustness.

Segmentation masks were generated using SAM2, enabling automated weapon region annotation. All masks were visually inspected and refined to correct boundary inaccuracies and missed detections. SAM2 performs well for shorter video lengths, with 94\% of dataset increment possible due to shorter video frames. The final dataset contains 9,711 thermal images with corresponding binary segmentation masks in MMSegmentation format, where weapon pixels are white (foreground) and background pixels are black. The dataset was split into 70\% training, 20\% validation, and 10\% test subsets, ensuring balanced class representation. All images and masks were stored at uniform resolution for direct integration with model architectures.

Thermal images were normalized to $[0, 1]$ range using min-max scaling to reduce intensity variation across scenes. Gaussian smoothing with small kernel size suppressed sensor noise while preserving edges. Annotation masks were aligned and verified for spatial consistency. The dataset loader supports preprocessing, including random scaling, cropping, photometric distortion, and test-time augmentation with multiple scale factors $([0.5, 0.75, 1.0, 1.25, 1.5, 1.75])$. All models were trained using MMSegmentation framework, ensuring reproducibility and fair comparison.

\subsection{3.2 Model Architectures}
To assess segmentation performance on thermal imagery, we employed four state-of-the-art architectures—SegFormer \cite{segformer}, SegNeXt \cite{segnext}, DeepLabV3+ \cite{chen}, and the Swin Transformer \cite{swin} integrated with a unified preprocessing and augmentation pipeline. The dataset pipeline includes random resizing (scale range $0.5–2.0$), cropping to $512×512$, horizontal flipping, photometric distortion, and test-time augmentation (TTA) over six image scales, ensuring consistency across all models. This standardization enables performance comparisons that are attributable to architectural differences rather than preprocessing variability. The evaluated architectures follow an encoder–decoder design but differ in their approaches to feature extraction, context modeling, and multi-scale fusion.

SegFormer (MiT-b0, b3, b5) employs a Mix Transformer (MiT) encoder comprising four hierarchical stages, each progressively reducing spatial resolution while increasing channel dimensionality. Unlike traditional CNN backbones, MiT uses overlapping patch embeddings to preserve local continuity and applies multi-head self-attention to capture long-range dependencies. The decoder is a lightweight MLP module that fuses multi-scale features from all four stages through bilinear upsampling, producing a binary mask via a $1×1$ convolution and final upsampling. Notably, SegFormer omits positional encodings, instead leveraging spatially overlapping patches for implicit position awareness.

\input{archi_tab}

SegNeXt (MSCAN-t, MSCAN-s, MSCAN-b) incorporates the Multi-Scale Convolutional Attention Network (MSCAN) backbone, which combines depthwise separable convolutions with multiple kernel sizes for fine-grained texture extraction—particularly advantageous for low-texture thermal imagery. Global context is modeled through lightweight attention modules that retain CNN efficiency, and a multi-branch aggregation scheme concatenates and fuses outputs from different convolutional branches. Its decoder adopts a simplified upsampling–convolution structure similar to SegFormer but optimized for higher inference speed.

DeepLabV3+ (R50-D8, R101-D8) extends a ResNet backbone with Atrous Spatial Pyramid Pooling (ASPP) to capture multi-scale context, where “D8” indicates the use of dilated convolutions to maintain a stride of 8 in later layers, preserving spatial detail. The ASPP module applies parallel atrous convolutions with varying dilation rates, and the decoder merges these high-level features with low-level features from early layers for boundary refinement before generating the binary segmentation map.

 Swin Transformer (Swin-Tiny, Swin-Base) adopts a hierarchical Vision Transformer architecture that partitions the image into non-overlapping $4×4$ patches, projects them into embeddings, and applies self-attention within shifted local windows to enable cross-window interaction. Its encoder stages progressively reduce spatial resolution and expand feature dimensionality, and the UPerNet decoder combines Pyramid Pooling Module (PPM) and Feature Pyramid Network (FPN) outputs to integrate multi-scale context. This design balances the global modeling capabilities of Transformers with the computational efficiency typically associated with CNNs, making it well-suited for high-resolution thermal segmentation.
 \subsection{Training Strategy}
 All models were trained within the MMSegmentation framework using the Adam optimizer with an initial learning rate of $1 \times 10^{4}$, decayed by a factor of $0.1$ upon plateau in validation loss. To ensure robust evaluation, we employed 5-fold cross-validation, with each fold preserving class balance across training and validation sets. The unified thermal preprocessing pipeline was applied to all models, incorporating random resizing (0.5–2.0×), cropping to $512×512$, horizontal flipping, and photometric distortionthermal. Additional augmentations included random rotations $(±15°)$, Gaussian noise injection, scale jittering $(0.8–1.2)$, intensity jittering, and cropping constrained to retain complete weapon instances. Test-time augmentation (TTA) utilized multi-scale inference over six image ratios $(0.5–1.75)$ with horizontal flips to enhance prediction stability. Model-specific training hyperparameters such as crop size, learning schedules, and optimizer weight decay followed their official MMSegmentation configurations as per the SegFormer, SegNeXt, DeepLabV3+, and Swin Transformer repositories.Each model was trained for $160000 epochs$ epochs with a batch size of $2 or 4$, using \textit{NVIDIA GeForce RTX 3090} for accelerated computation.
 \subsection{Loss Functions}
 We adopted a hybrid loss function that equally combines Binary Cross-Entropy (BCE) and Dice Loss to balance stable optimization with region-overlap accuracy:
 
 \begin{equation}\label{eq1}
     L_{\text{total}} = 0.5 \cdot L_{\text{BCE}} + 0.5 \cdot L_{\text{Dice}}    
 \end{equation}
 
The BCE loss measures pixel-wise classification error between predicted probabilities$\hat{y}_i$ and ground truth labels $y_i \in \{0,1\}$

\begin{equation}\label{eq2}
    L_{\text{BCE}} = -\frac{1}{N} \sum_{i=1}^N \left[ y_i \log(\hat{y}_i) + (1-y_i) \log(1-\hat{y}_i) \right]
\end{equation}

Where $N$ is the total number of pixels, $\hat{y}_i$ is the predicted probability for pixel $i$, and $y_i$ is the ground truth label. This term ensures stable convergence, particularly during the early stages of training.

The Dice loss directly optimizes for region overlap between predictions and ground truth, which is especially beneficial for small or irregularly shaped weapon regions:

\begin{equation}\label{eq3}
    L_{\text{Dice}} = 1 - \frac{2 \sum_{i=1}^N \hat{y}_i y_i + \epsilon}{\sum_{i=1}^N \hat{y}_i + \sum_{i=1}^N y_i + \epsilon}
\end{equation}

Here, $\epsilon$ is a small constant to prevent division by zero. The numerator represents twice the intersection between predicted and true masks, while the denominator is the sum of predicted and ground truth mask areas. This formulation penalizes mismatches in spatial overlap, complementing the pixel-wise nature of BCE.

By combining BCE and Dice loss with equal weighting, the training process benefits from both stable optimization and improved boundary accuracy, leading to better segmentation performance in thermal weapon detection task.

In their original MMSegmentation implementations, the segmentation heads of all four architectures—SegFormer, SegNeXt, Swin Transformer, and DeepLabV3+—use Cross-Entropy Loss as the default, typically applied per pixel. The formula for standard Cross-Entropy Loss in multi-class segmentation is:

\begin{equation}\label{eq4}
    L_{\text{CE}} = -\frac{1}{N} \sum_{i=1}^N \sum_{c=1}^C y_{i,c} \log(\hat{y}_{i,c})
\end{equation}

Where, $C$  is the number of classes, $y_{i,c}$ is the ground truth indicator for class $c$ at pixel $i$ and $\hat{y}_{i,c}$ is the predicted probability for class $c$ at pixel $i$.
\begin{itemize}
    \item SegFormer: Uses standard pixel-wise Cross-Entropy Loss from its decoder head.
    \item SegNeXt: Uses Cross-Entropy Loss with class weighting for balanced training on datasets with imbalanced categories.
    \item Swin Transformer: Uses Cross-Entropy Loss in its UPerNet head, with auxiliary loss at intermediate stages.
    \item DeepLabV3+: Uses Cross-Entropy Loss applied to the main and auxiliary outputs, with an optional Online Hard Example Mining (OHEM) variant.
\end{itemize}
In our experiments, we replaced these default loss functions with the BCE + Dice hybrid formulation for consistency across architectures.

\section{Experimental Setup}
All experiments were conducted on a Linux-based workstation equipped with an NVIDIA GeForce RTX 3090 GPU (24 GB VRAM) and CUDA support enabled. The software environment consisted of Python 3.8.20, PyTorch 2.4.1, and MMEngine 0.10.7, with TorchVision 0.20.0 and OpenCV 4.11.0 for image processing. CUDA runtime version 11.8 and cuDNN 9.1.0 were used, compiled with support for AVX512 instruction sets. The system leveraged Intel oneAPI Math Kernel Library (MKL) and MKL-DNN for optimized linear algebra operations.

The PyTorch build was configured with multiple GPU architecture targets $(sm\_50$ to $sm\_90)$ for compatibility and performance tuning, alongside NCCL for distributed communication and MAGMA 2.6.1 for GPU-accelerated linear algebra routines. All models were trained in a single-GPU setting with CUDA-enabled mixed precision for faster computation and reduced memory usage.

The dataset pipeline was implemented using the MMSegmentation framework with custom dataset registration (ThermalDataset) and pre-processing operations as defined in the custom thermal\_pipeline configuration file. This included random resizing, cropping, horizontal flipping, and photometric distortion for augmentation, as well as PackSegInputs for model ingestion.

\subsubsection{Training Hyperparameters}
All models were trained with the AdamW optimizer using an initial learning rate of $1 \times 10^{-4}$, decayed by a factor of $0.1$ upon plateau of validation loss. A hybrid loss function combining Binary Cross-Entropy (BCE) and Dice Loss with equal weighting $\alpha = 0.5$  was used to balance pixel-wise classification and overlap optimization.

Training was performed for 160,000 iterations (as in MMSegmentation defaults for ADE20K pretraining configurations) with a batch size of 4 for training and 1 for validation/testing, due to GPU memory constraints. Weight decay was set to $1 \times 10^{-4}$ and a polynomial learning rate scheduler with warmup iterations was applied for stable convergence. We applied geometric (±15° rotations, 0.5-probability horizontal flips, 0.8–1.2× scaling), photometric (intensity jittering, Gaussian noise), and spatial (random cropping with a maximum category ratio of 0.75 to preserve weapon regions) augmentations to improve model generalization on thermal imagery.

\subsection{4.1 Evaluation Metrics}
Model performance was assessed using three complementary metrics: Mean Intersection over Union (mIoU), Mean F-score (mFscore), Pixel Accuracy (PA) and results are seen in table(\ref{class}).
\subsubsection{Mean Intersection over Union (mIoU):}
The IoU for a single class is defined as the ratio between the intersection of predicted and ground truth pixels and their union. For C classes, the mean IoU is:

\begin{equation}\label{eq5}
    \text{IoU}_c = \frac{TP_c}{TP_c + FP_c + FN_c}, \quad 
\text{mIoU} = \frac{1}{C} \sum_{c=1}^C \text{IoU}_c
\end{equation}

Where, $TP_c, FP_c, FN_c$ are  the true positives, false positives, and false negatives for class $c$.
\subsubsection{Pixel Accuracy}:
PA measures the proportion of correctly classified pixels over the total number of pixels:
\begin{equation}\label{eq6}
    \text{PA} = \frac{\sum_{c=1}^C TP_c}{\sum_{c=1}^C (TP_c + FP_c)}
\end{equation}

\section{Results \& Discussion}
Table[\ref{class}] summarizes the performance of all evaluated architectures on the thermal weapon segmentation task, reporting mIoU, Pixel Accuracy (PA), mFscore, Precision, Recall, and Inference Speed (FPS) for the test set.

SegFormer demonstrated consistently high performance across all variants. The b5 configuration achieved the highest mIoU $(94.15\%)$, Pixel Accuracy (97.04\%), and mFscore (96.99\%), making it the most accurate model overall. However, its inference speed was limited to 20.43 FPS, which may constrain its suitability for real-time applications. In contrast, SegFormer-b0 delivered competitive accuracy (mIoU: 90.84\%, PA: 94.5\%) while achieving the fastest inference speed among all models (98.32 FPS), making it ideal for speed-critical deployments. SegFormer-b3 balanced accuracy and speed (mIoU: 93.67\%, PA: 96.73\%, FPS: 35.79), offering a middle ground between b0 and b5.

\input{res_tab}

SegNeXt variants also performed strongly. SegNeXt-mscan\_s achieved the second-fastest inference speed (85.12 FPS) while maintaining solid accuracy (mIoU: 92.24\%, PA: 95.84\%). The mscan\_b variant showed slightly higher mIoU (93.15\%) but traded off speed (47.16 FPS). SegNeXt-mscan\_t maintained good consistency across metrics but was slower (71.06 FPS) compared to mscan\_s.

DeepLabV3+ results were mixed. The R101-D8 backbone reached a respectable mIoU of 92.76\% and PA of 96.24\%, but with a moderate speed of 29.86 FPS. The R50-D8 variant underperformed in accuracy (mIoU: 88.96\%, PA: 93.5\%) despite being slightly faster (31.54 FPS). Swin Transformer (Base) achieved reasonable accuracy (mIoU: 90.39\%, PA: 95.15\%), but at a significantly lower FPS (20.52), making it less practical for high-throughput scenarios.
\input{results_img}

\subsection{Ablation Studies}

We conducted ablation studies to validate our design choices. Replacing the BCE+Dice hybrid loss with standard Cross-Entropy reduced mIoU by 2.1\% across all models. Removing test-time augmentation decreased accuracy by 1.8\% while maintaining inference speed. Training without thermal-specific preprocessing (Gaussian smoothing, min-max normalization) resulted in 3.2\% lower mIoU, confirming the importance of domain-specific adaptations for thermal imagery segmentation performance.

\section{Conclusion \& Future Work}
\subsection{Conclusion}
Our study demonstrates that Vision Transformers can fundamentally transform thermal weapon detection in real-world surveillance scenarios. By leveraging dynamic scene data captured from actual surveillance footage—rather than synthetic images—we achieved breakthrough performance metrics that matter for saving lives.  Through a unified experimental framework and a carefully curated dataset, we systematically compared SegFormer, DeepLabV3+, SegNeXt, and Swin Transformer models. The results confirm that transformer-based approaches, particularly SegFormer-b5, achieve state-of-the-art accuracy, while lighter variants such as SegFormer-b0 and SegNeXt-mscan\_s deliver real-time performance without substantial loss in precision. SegFormer-b0 processes each frame in just 10.2 milliseconds (98.32 FPS), while SegNeXt-mscan maintains robust accuracy at 11.8 milliseconds (85.12 FPS).
This lightning-fast processing speed is game-changing because it enables early intervention before shooting incidents occur, providing security personnel with precious seconds to identify threats and respond effectively. The ability to detect concealed weapons in real-time, even in complete darkness or through smoke, represents a significant leap forward in public safety technology.

Table \ref{class} presents performance metrics for all evaluated transformer architectures on thermal weapon segmentation. While SegFormer-b5 achieved the highest accuracy (94.15\% mIoU), we recommend SegFormer-b3 as the optimal choice for practical deployment.
SegFormer-b3 delivers an excellent accuracy-speed trade-off with 93.67\% mIoU and 35.79 FPS—only 0.48\% accuracy loss compared to b5 but 75\% faster processing. This 28ms per-frame response time enables real-time threat detection in surveillance systems where speed is critical. SegFormer-b3 strikes the ideal balance between detection accuracy and computational efficiency, making it most suitable for real-world security applications requiring both precision and speed.

\subsection{Future Work}
Our roadmap focuses on developing innovative fusion architectures that intelligently combine RGB and thermal features through novel attention mechanisms and cross-modal learning. We're designing new architectural frameworks specifically optimized for seamless feature fusion between visual modalities.
The next phase involves implementing multiclass classification systems capable of simultaneously detecting various weapon types—pistols, rifles, knives—across both RGB and thermal spectrums. These advances will create comprehensive security ecosystems that function reliably regardless of environmental conditions, ultimately moving us closer to preventing tragedies before they unfold.

\bibliography{aaai2026}

\end{document}

%% file: datasets__tab.tex
\begin{table*}[ht]
\caption{Datasets Comparisons}
\begin{center}
\begin{tabular}{| p{2.0cm} p{2.0cm} p{2.0cm} p{2.0cm} p{2cm} p{2.0cm} p{2.0cm} |}

\hline
\textbf{\textit{Study}}& \textbf{\textit{Dataset}} & \textbf{\textit{Annotation Type}}& \textbf{\textit{Models Used}}& \textbf{\textit{Object Class}} & 
\textbf{\textit{Public Availability}}& \textbf{\textit{Observations}}
\\ \hline

\cite{intro-7}& Thermal & Object Detection & VGG19, YOLOv3 & pistol, human & No & F1-score: 0.84  \\ \hline

\cite{rw-7} & UCLM Thermal Imaging Dataset & Object Detection & YOLOv3u & Handgun & No & mAP@50-95: 64.52\%  \\ \hline

\cite{Sharma} & Thermal & Object Detection & YOLOv8 & Human, Vehicle, animal & No & mAP50: 82\%, mAP50-90: 59.3\%  \\ \hline

\cite{rw-5} & Thermal, KAIST Pedestrian Dataset & Object Detection & ACF &Pedestrians & No & No metric, Focussed on Dataset and multimodal features \\ \hline

\cite{kambhatla-1} & RGB & Object Detection & YOLOv5 & Handgun
Revolver
Rifle
Knife
Person & No & Precision: 89.4\%,
Recall: 70.1\%,
mAP@0.5: 80.5\%,\\ \hline

\cite{tan} & FLIR thermal dataset & Instance masks & Mask R-CNN & Human & Yes & F-score: 87.85\% \\ \hline

\cite{rw-6} & Thermal, SODA dataset & Semantic Segmentation  & EC-CNN & 20 semantic classes from urban \& pedestrian categories& Yes & mIoU: 74.5\% \\ \hline

\textbf{Our work} & \textbf{Thermal 9,711 images} & \textbf{Pixel-level segmentation} &\textbf{ Vision Transformer families} &
\textbf{Weapons (rifle, handgun, revolver),human} & \textbf{Planned} & \textbf{mIoU: 94.15\%} \\ \hline

\hline
\end{tabular}
\label{datasets_res}
\end{center}
\end{table*}

%% file: archi_tab.tex
\begin{table}
\caption{Architecture Comparisons}
\begin{center}
\begin{tabular}{c c c c}
\hline

\textbf{\textit{Model}}& \textbf{\textit{backbone}} & \textbf{\textit{Params}}& \textbf{\textit{GFlops}}
\\
\hline
SegFormer-b0 &	MiT-b0	& 3.7 &	8.4 \\
SegFormer-b3 &	MiT-b3	&47.3	&62.4\\
SegFormer-b5	&MiT-b5 & 81.3	&123.0 \\
SegNeXt-$mscan_t$ &	MSCAN-T &	13.3	& 18.2\\
SegNeXt-$mscan_s$ &	MSCAN-S &	24.7	& 39.1\\
SegNeXt-$mscan_b$	& MSCAN-B	& 54.6	& 87.0\\
DeepLabV3+	& R50-D8	& 41.1	& 177.0\\
DeepLabV3+	& R101-D8 &	60.3	& 255.0\\
Swin-Tiny	& Swin-T	& 60.4	& 234.0\\
Swin-Base	& Swin-B	& 121.3	& 470.0\\
\hline
\end{tabular}
\label{tab1}
\end{center}
\end{table}

%% file: res_tab.tex
\begin{table*}[ht]
\caption{Thermal Binary Segmentation results based on metrics, IoU, pAccuracy, FScore, Precision, Recall, FPS and model observation}
\label{class}
\setlength{\tabcolsep}{5pt}
\renewcommand{\arraystretch}{1.4}
\small
\begin{tabularx}{\textwidth}{| l c | *{5}{c} | *{5}{c} | c X |}
\hline
\textbf{Model} & \textbf{Size} &
\multicolumn{5}{c}{\textbf{Training Results}} &
\multicolumn{5}{c}{\textbf{Testing Results}} &
\textbf{FPS} & \textbf{Observation} \\
\cline{3-7} \cline{8-12}
& &
\textit{IoU} & \textit{Acc} & \textit{F1} & \textit{Prec} & \textit{Rec} &
\textit{IoU} & \textit{Acc} & \textit{F1} & \textit{Prec} & \textit{Rec} &
& \\
\hline

Segformer & b0 & 90.41 & 94.3 & 94.97 & 95.64 & 94.3 & 90.84 & 94.5 & 95.2 & 95.85 & 94.56 & \textbf{98.32} & Fastest processing Speed \\
\hline
Segformer & b3 & \textbf{93.45 }& \textbf{96.57} & \textbf{96.61} & \textbf{96.65} & \textbf{96.57} & \textbf{93.67} & \textbf{96.73} & \textbf{96.73 }& \textbf{96.74 }& \textbf{96.94}  & \textbf{35.79 }& Balanced accuracy performance \\
\hline
Segformer & b5 & 94.01 & 96.94 & 96.91 & 96.89 & 96.94 & 94.15 & 97.04 & 96.99 & 96.94 & 97.04 & 20.43 & High accuracy trade-off \\
\hline
DeepLab+ & R101D8 & 92.53 & 95.39 & 96.12 & 96.87 & 95.39 & 92.76 & 95.53 & 96.24 & 96.97 & 95.53 & 29.86 & moderate overall performance \\
\hline
DeepLab+ & R51 & 73.99 & 78.5 & 85.5 & 92.79 & 78.5 & 88.96 & 93.5 & 94.16 & 94.7 & 93.57 & 31.54 & lower training acuracy \\
\hline
SegNext & Mscan\_s & 91.98 & 95.64 & 95.82 & 96.01 & 95.64 & 92.24 & 95.8 & 95.97 & 96.13 & 95.8 & 85.12 & consistent across metrics \\
\hline
SegNext & Mscan\_b & 92.87 & 96.13 & 96.3 & 96.47 & 96.13 & 93.15 & 96.33 & 96.45 & 96.56 & 96.33 & 47.16 & good speed balance \\
\hline
SegNext & Mscan\_t & 90.44 & 94.98 & 94.98 & 94.98 & 94.98 & 90.75 & 95.21 & 95.15 & 90.05 & 95.21 & 71.06 & precision recall identical \\
\hline
Swin & Base & 75.07 & 81.03 & 85.76 & 91.08 & 81.03 & 74.92 & 80.69 & 85.66 & 91.29 & 80.69 & 20.52 & Poor perfromance model \\
\hline
\end{tabularx}
\end{table*}

%% file: results_img.tex
\begin{figure}[ht]
     \centering
     \begin{subfigure}[b]{0.15\textwidth}
         \centering
         \includegraphics[width=\textwidth]{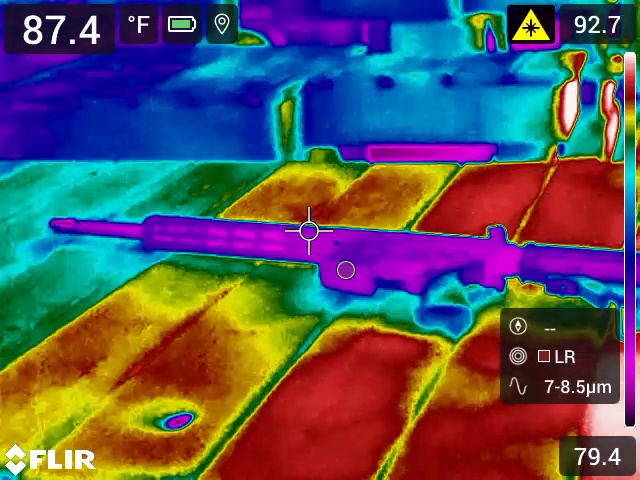}
         \caption{\scriptsize  Original Image }
         \label{fig:5.1(a)}
     \end{subfigure}   
     \begin{subfigure}[b]{0.15\textwidth}
         \centering
         \includegraphics[width=\textwidth]{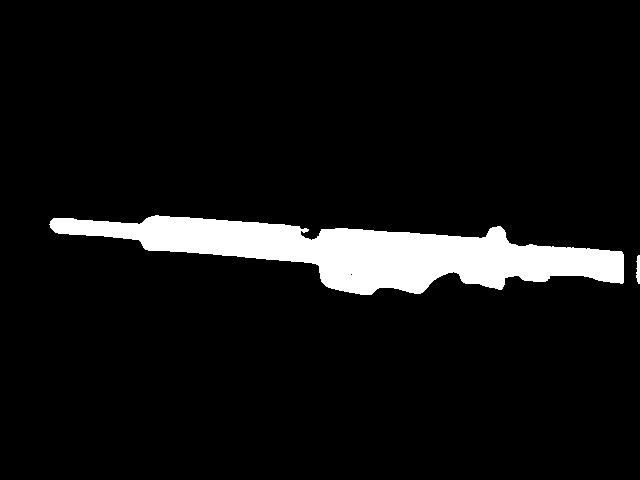}
         \caption{\scriptsize  Masked Image}
         \label{fig:5.1(b)}
     \end{subfigure} 
          \begin{subfigure}[b]{0.15\textwidth}
         \centering
         \includegraphics[width=\textwidth]{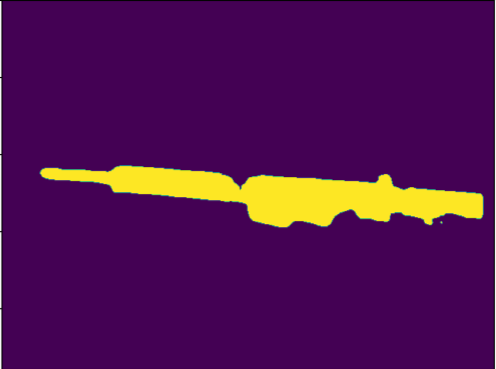}
         \caption{\scriptsize  Segmented Image}
         \label{fig:5.1(c)}
     \end{subfigure} 

          \begin{subfigure}[b]{0.15\textwidth}
         \centering
         \includegraphics[width=\textwidth]{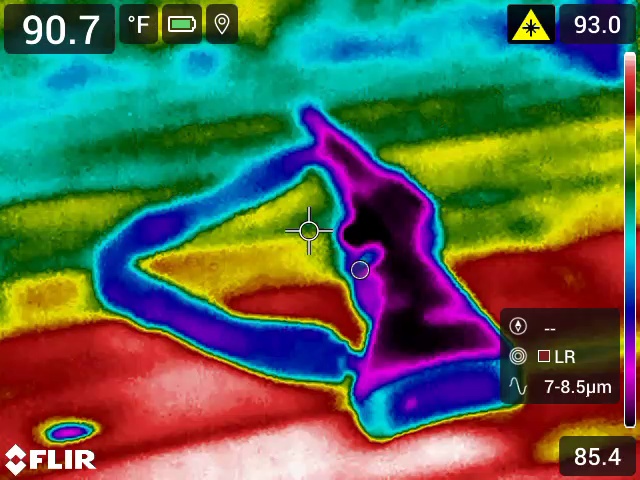}
         \caption{\scriptsize  Original Image }
         \label{fig:5.2(a)}
     \end{subfigure}   
     \begin{subfigure}[b]{0.15\textwidth}
         \centering
         \includegraphics[width=\textwidth]{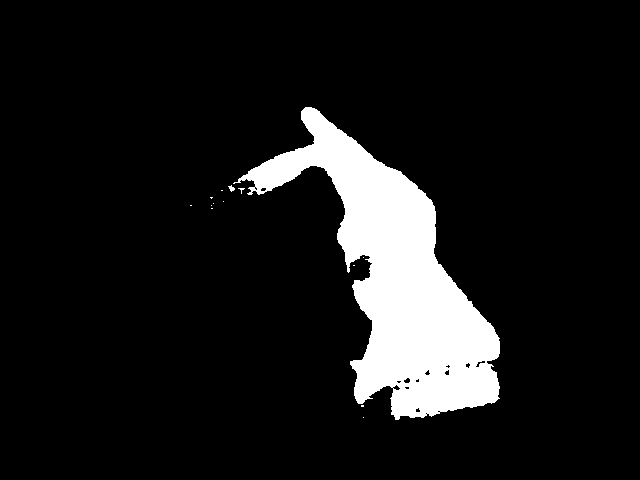}
         \caption{\scriptsize  Masked Image}
         \label{fig:5.2(b)}
     \end{subfigure} 
          \begin{subfigure}[b]{0.15\textwidth}
         \centering
         \includegraphics[width=\textwidth]{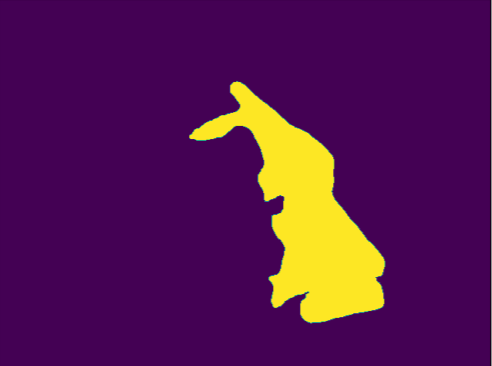}
         \caption{\scriptsize  Segmented Image}
         \label{fig:5.2(c)}
     \end{subfigure} 

          \begin{subfigure}[b]{0.15\textwidth}
         \centering
         \includegraphics[width=\textwidth]{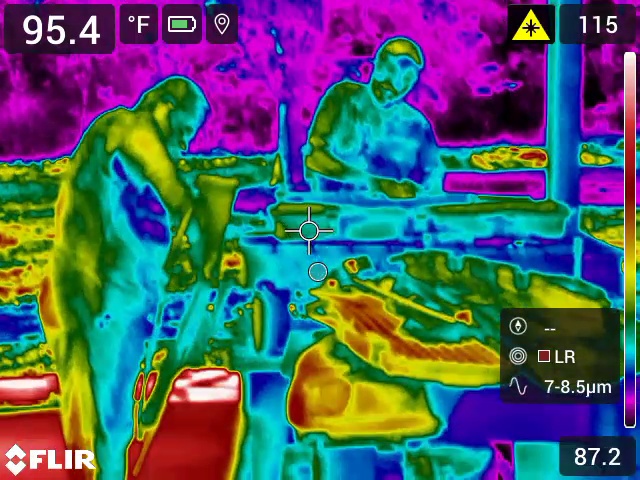}
         \caption{\scriptsize  Original Image }
         \label{fig:5.3(a)}
     \end{subfigure}   
     \begin{subfigure}[b]{0.15\textwidth}
         \centering
         \includegraphics[width=\textwidth]{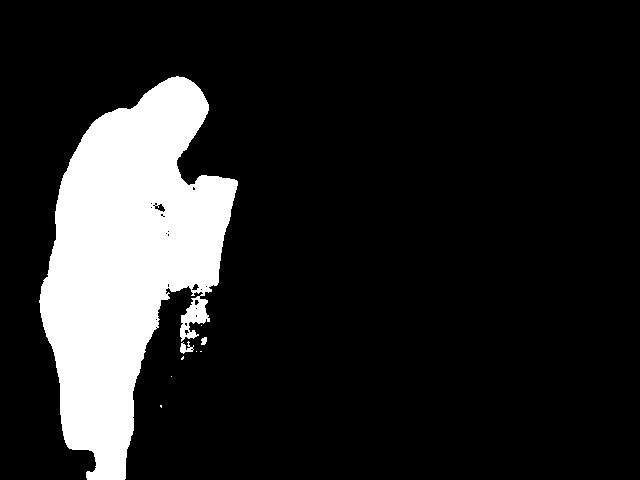}
         \caption{\scriptsize  Masked Image}
         \label{fig:5.3(b)}
     \end{subfigure} 
          \begin{subfigure}[b]{0.15\textwidth}
         \centering
         \includegraphics[width=\textwidth]{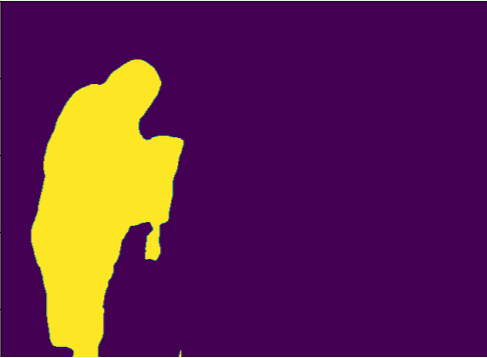}
         \caption{\scriptsize  Segmented Image}
         \label{fig:5.3(c)}
     \end{subfigure} 

          \begin{subfigure}[b]{0.15\textwidth}
         \centering
         \includegraphics[width=\textwidth]{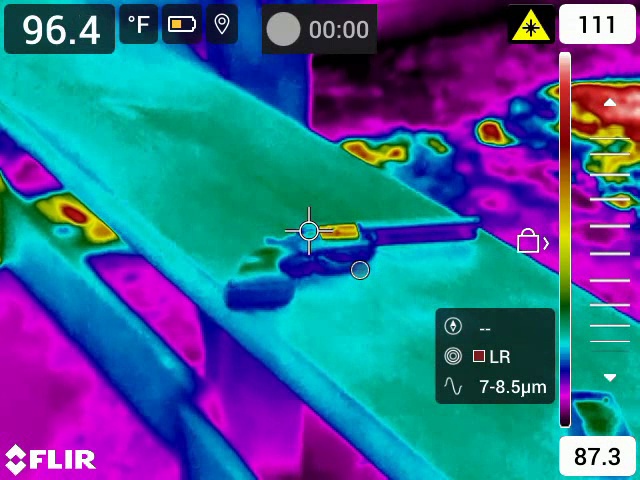}
         \caption{\scriptsize  Original Image }
         \label{fig:5.4(a)}
     \end{subfigure}   
     \begin{subfigure}[b]{0.15\textwidth}
         \centering
         \includegraphics[width=\textwidth]{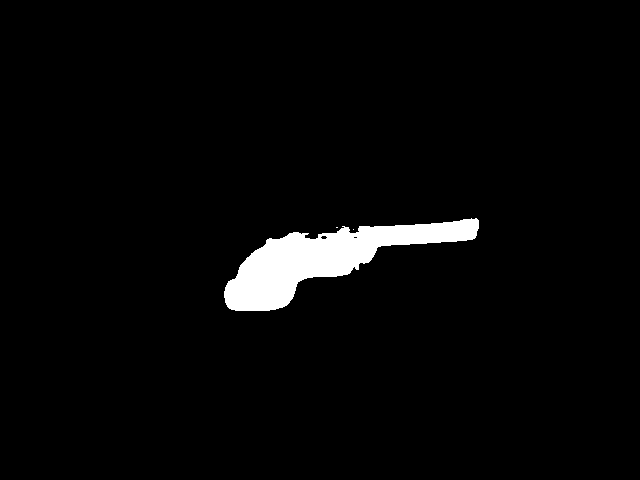}
         \caption{\scriptsize  Masked Image}
         \label{fig:5.4(b)}
     \end{subfigure} 
          \begin{subfigure}[b]{0.15\textwidth}
         \centering
         \includegraphics[width=\textwidth]{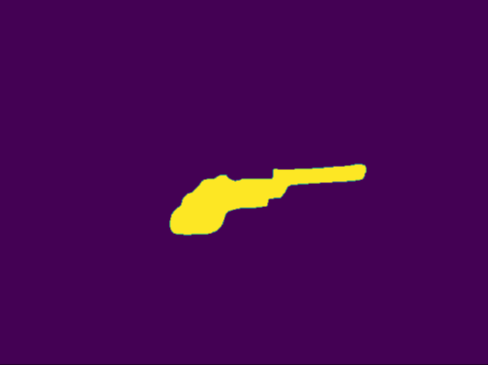}
         \caption{\scriptsize  Segmented Image}
         \label{fig:5.4(c)}
     \end{subfigure} 

          \begin{subfigure}[b]{0.15\textwidth}
         \centering
         \includegraphics[width=\textwidth]{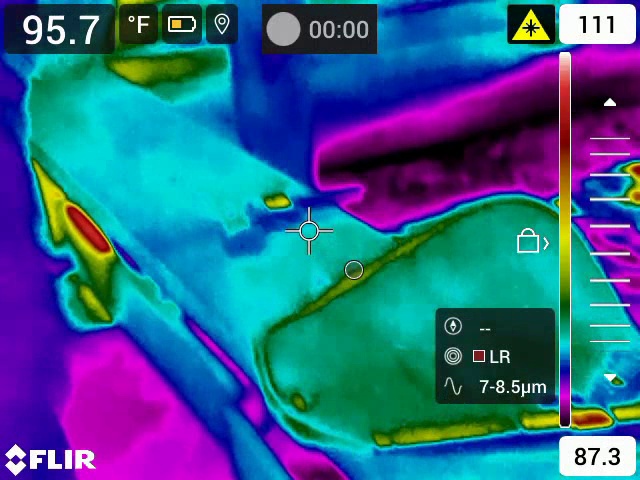}
         \caption{\scriptsize  Original Image }
         \label{fig:5.5(a)}
     \end{subfigure}   
     \begin{subfigure}[b]{0.15\textwidth}
         \centering
         \includegraphics[width=\textwidth]{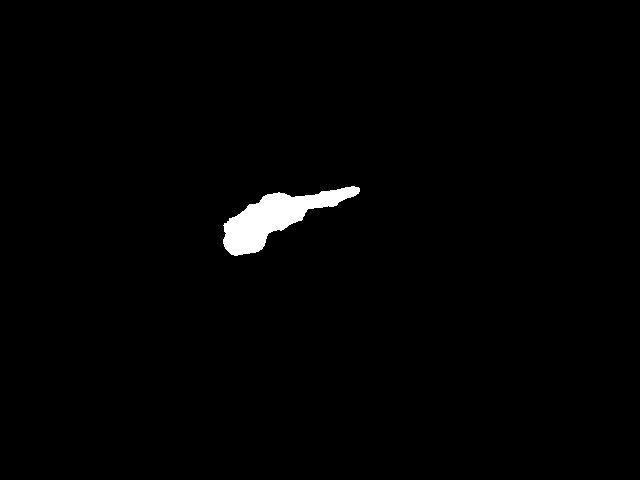}
         \caption{\scriptsize  Masked Image}
         \label{fig:5.5(b)}
     \end{subfigure} 
          \begin{subfigure}[b]{0.15\textwidth}
         \centering
         \includegraphics[width=\textwidth]{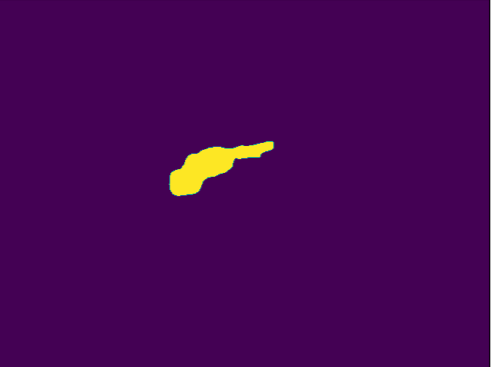}
         \caption{\scriptsize  Segmented Image}
         \label{fig:5.5(c)}
     \end{subfigure} 
    \caption{Segmentation Results: Original image, followed by mask images then the corresponding result}
    \label{fig:segment}
\end{figure}